\documentclass{article}

\usepackage{PRIMEarxiv}

\usepackage[utf8]{inputenc} 
\usepackage[T1]{fontenc}    
\usepackage{hyperref}       
\usepackage{url}            
\usepackage{booktabs}       
\usepackage{amsfonts}       
\usepackage{nicefrac}       
\usepackage{microtype}      
\usepackage{lipsum}
\usepackage{fancyhdr}       
\usepackage{graphicx}       
\graphicspath{{media/}}     

\title{Target Features Affect Visual Search, A Study of Eye Fixations}

\author{
  Manoosh Samiei \\
  Center for Intelligent Machines\\
  McGill University \\
  Montreal, Quebec, Canada\\
  \texttt{manoosh.samiei@mail.mcgill.ca} \\
  \And
  James J. Clark \\
  Center for Intelligent Machines\\
  McGill University \\
  Montreal, Quebec, Canada\\
  \texttt{james.clark1@mcgill.ca} \\
}

\begin{document}
\maketitle

\begin{abstract}

Visual Search is referred to the task of finding a target object among a set of distracting objects in a visual display. In this paper, based on an independent analysis of the COCO-Search18 dataset, we investigate how the performance of human participants during visual search is affected by different parameters such as the size and eccentricity of the target object.  We also study the correlation between the error rate of participants and search performance. Our studies show that a bigger and more eccentric target is found faster with fewer number of fixations. 
Our code for the graphics are publicly available at: \url{https://github.com/ManooshSamiei/COCOSearch18_Analysis} 
\end{abstract}


\section{Introduction}
Visual Search provides the organisms with many survival benefits such as finding food and avoiding danger. Organisms that are equipped with a foveated visual system, i.e. having high visual acuity within ~1.5 degree of the fovea, move their eyes to perceive different parts of the scene with high resolution. The rapid eye movement that moves the gaze from one location to another location is referred to as saccade, and maintaining of the eyes on one location for a certain period of time between the saccades is referred to as fixation. As the eyes move rapidly during saccades the image received by the fovea is of low acuity; thus, the detailed processing of visual environment is mostly done during fixations. Depending on the difficulty of a search task, humans find a target object in the scene by a certain number of fixations. 
In this paper, we investigate the effect of the eccentricity (positioning in the center of the image), and the size of the target object on the search performance. In our analyses, all objects except the target are referred to as distractors(non-target) objects. We define multiple parameters as a measure of search performance. These parameters are: average reaction time, average non-target fixation duration time, average target fixation duration time, average total number of fixations, and average number of fixations on the target object. In all of these parameters, except the average reaction time, we exclude the initial center fixation. The initial center fixation refers to a dot appearing at the center of the screen, at which participants were asked to fixate, in the beginning of each trial. Furthermore, we study the correlation between these search parameters and the error rate of the participants.

\section{Dataset}

COCO-Search18 \cite{article} is a dataset that contains the eye fixation information of 10 participants while searching for each of 18 object categories in 6202 images. Among these images, 3101 do not contain the target object. In our analyses, we only use the 3101 images that contain the target. COCO-Search18 provides us with the fixation location, fixation duration, the total number of fixations, and reaction time, for each search trial. During data collection, participants were initially informed of the target category via a display showing the name of the target category along with sample images that exemplify that category. Also to avoid potential confusion negative examples were shown by placing a red X through the object that is not considered as a member of the target category. Participants then started a trial by pressing a button on a game-pad controller while looking at the fixation dot appearing at the center of the screen. afterwards, an image was displayed and the participant’s task was to quickly determine if the target is present in the image or not, by responding ‘yes’ or ‘no’ using the right or left triggers of a game-pad controller.  All image stimuli in the dataset have 1680 × 1050 pixel resolution. The 18 categories of target objects are: bottle, bowl, car, chair, clock, cup, fork, keyboard, knife, laptop, microwave, mouse, oven, potted plant, sink, stop sign, toilet, and TV; all of which appear in their natural context. 

\section{Analysis}

\subsection{Target Area}
\begin{figure*}[ht!]

\centering
\includegraphics[width=0.45\linewidth]{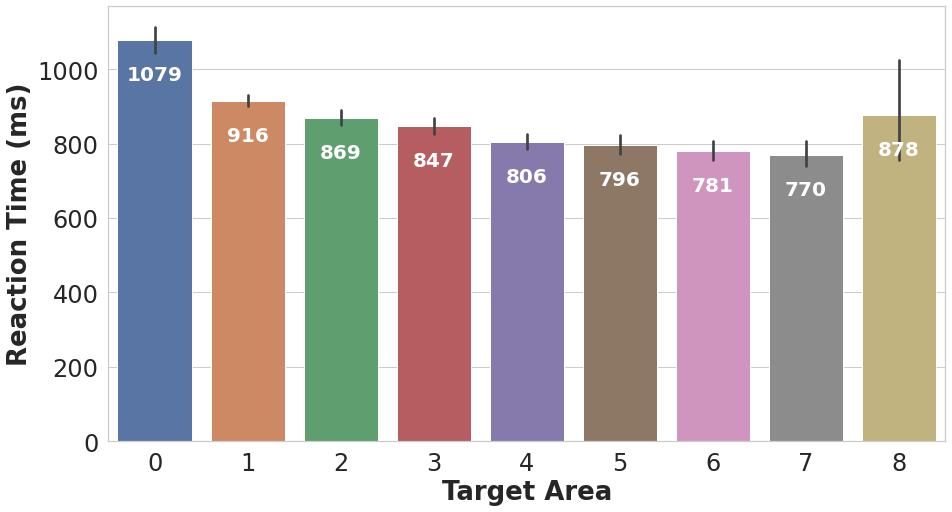}\label{fig:SGD_1_single}
\hfill
\includegraphics[width=0.45\linewidth]{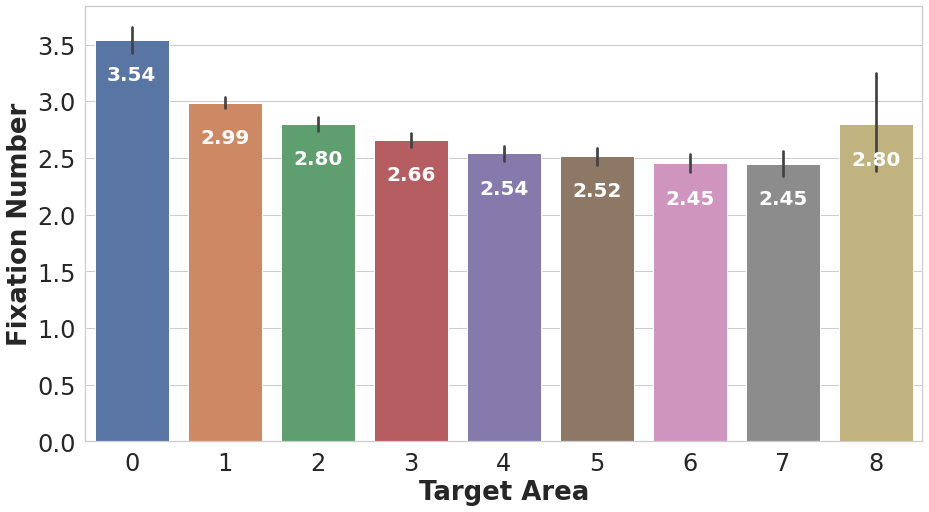}\label{fig:SGD_2_single}
\includegraphics[width=0.45\linewidth]{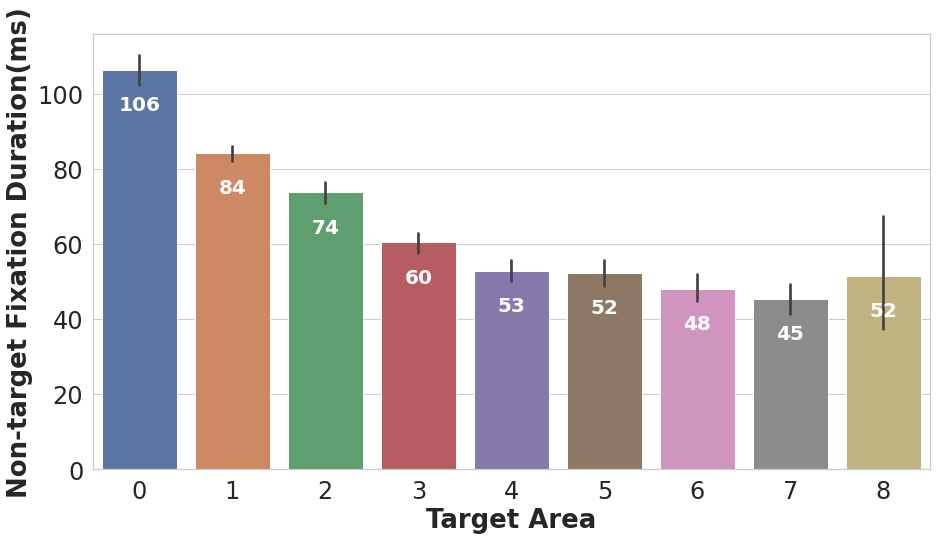}\label{fig:SGD_3_single}
\hfill
\includegraphics[width=0.45\linewidth]{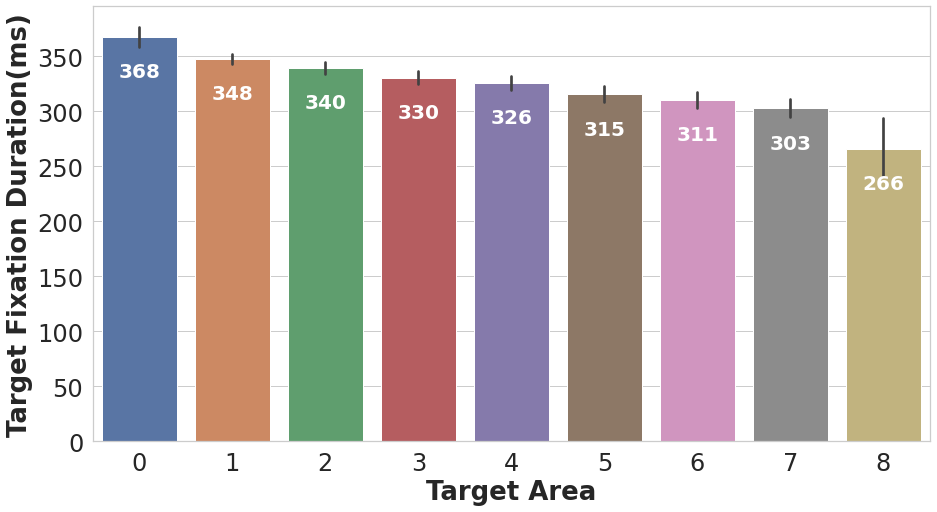}\label{fig:SGD_4_single}
\includegraphics[width=0.45\linewidth]{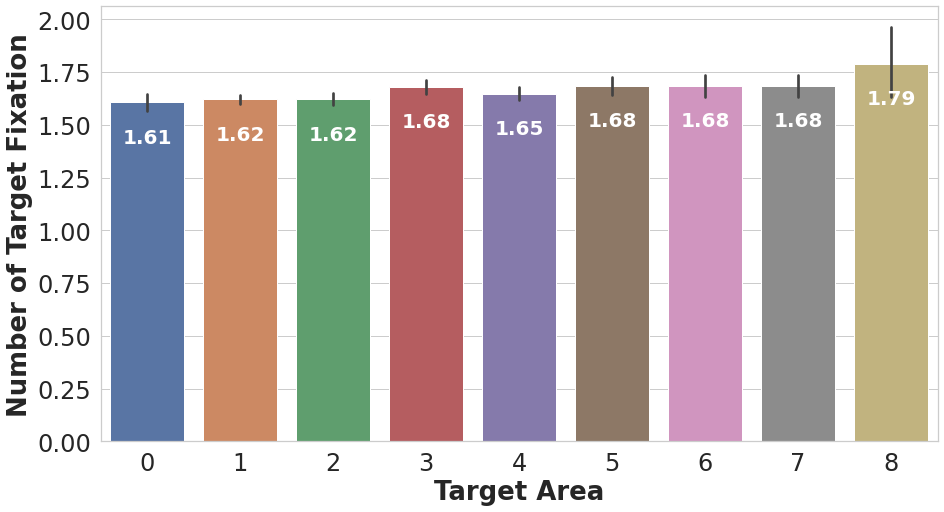}\label{fig:SGD_5_single}
\caption{Search performance versus target object size (area). (a) Average reaction time. (b) Average fixation number. (c) Average non-target fixation duration. (d) Average target fixation duration. (e) Average number of target fixation. }
\label{fig:area}
\end{figure*}

The characteristics of a target object have significant impact on search performance. One of these important characteristics is the size or area of the target object. We assume that the bigger the target, the higher the visibility. So the bigger targets should be found faster and with fewer fixations. We investigate this assumption with multiple figures, by plotting the average reaction time, average number of fixations (excluding the initial center fixation, including the target fixations), average non-target fixation duration time (excluding the initial center fixation), average target fixation duration time (excluding the initial center fixation), and average number of fixations on the target object versus the target area, in figure \ref{fig:area}. Target area is calculated by multiplying the width and height of the bounding box surrounding the target object. The calculated areas are then separated into bins, where every bin has a range of 2000 pixels. Thus, bin 0 covers 0-2000 pixel area, bin 1 covers 2000-4000 area, ..., and bin 8 covers 16000-18000 area. There is no target object with area bigger than 18000 pixels in images.

In figure \ref{fig:area}, we observe that the response time is higher for smaller targets and lower for bigger targets, excluding the last bin of the histogram which contains the biggest target size. The results of the last bin are less accurate as suggested by the error bars. This is because we have fewer cases where the target object area is between 16000-18000 pixels. We can thus rely on the rest of the bins to estimate the relationship between the variables. The number of fixations as well as distractor fixation duration time also decrease as the target area increases (probably due to pop-out effect, i.e. the target object stands out from the non-targets). The target duration time also decreases for bigger targets. This might be due to less focus needed to register bigger targets' information. The average number of fixations made on the target is relatively constant, with slightly more average fixations on bigger targets. This is intuitive; as the bigger the target, the higher the probability of fixating at it, since it occupies more space in the image.

\subsection{Target Eccentricity}

\begin{figure*}[ht!]
\centering
\includegraphics[width=0.45\linewidth]{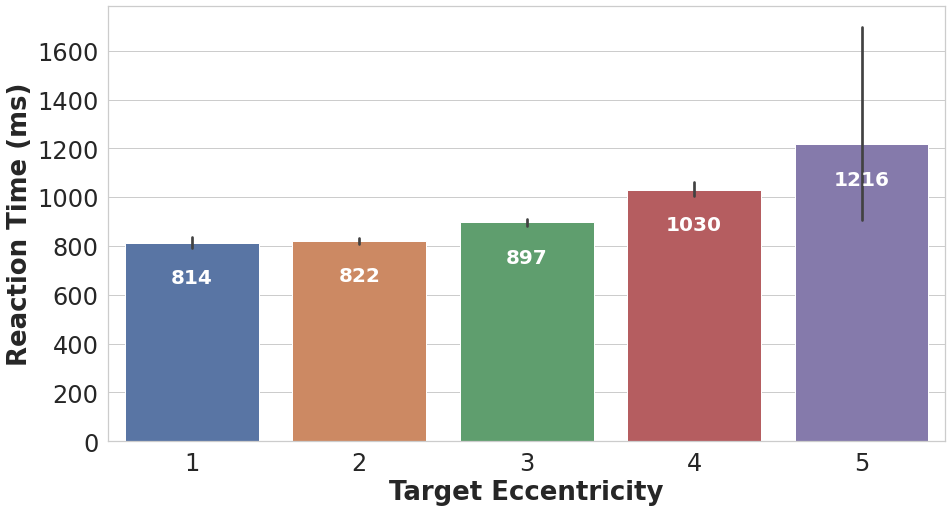}\label{fig:eccent_1_single}
\hfill
\includegraphics[width=0.45\linewidth]{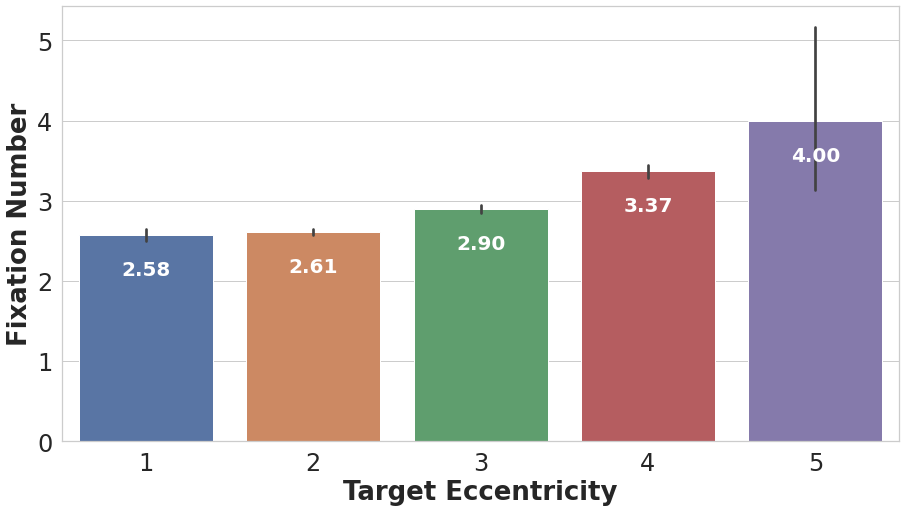}\label{fig:eccent_1_single}
\hfill
\centering
\includegraphics[width=0.45\linewidth]{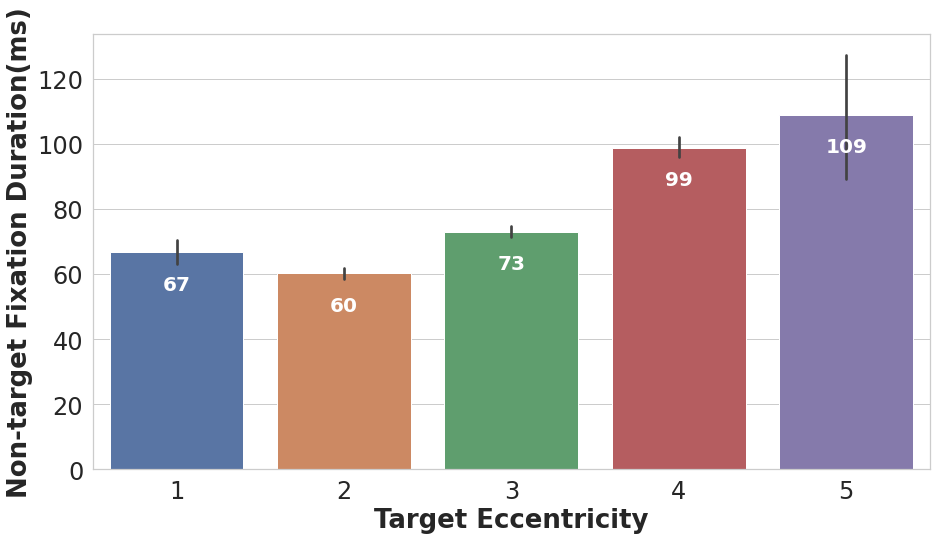}\label{fig:eccent_1_single}
\hfill
\centering
\includegraphics[width=0.45\linewidth]{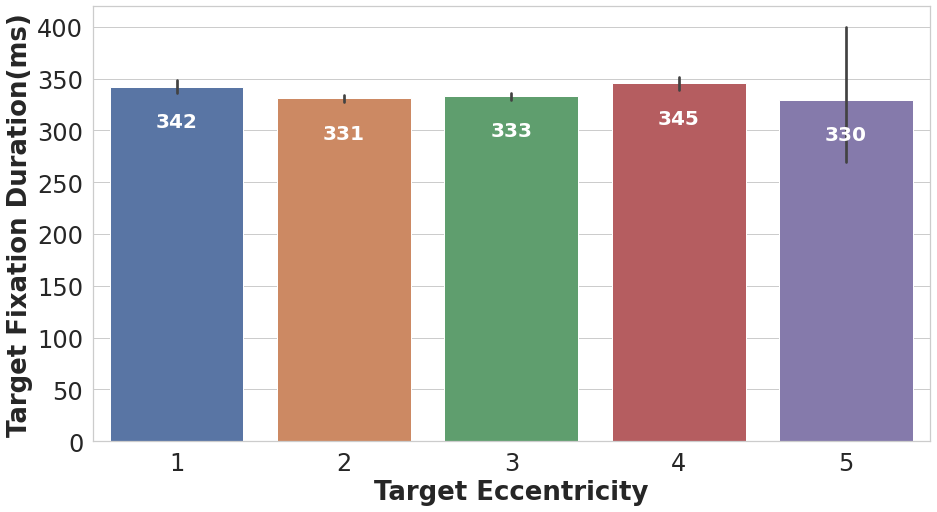}\label{fig:eccent_1_single}
\hfill
\centering
\includegraphics[width=0.45\linewidth]{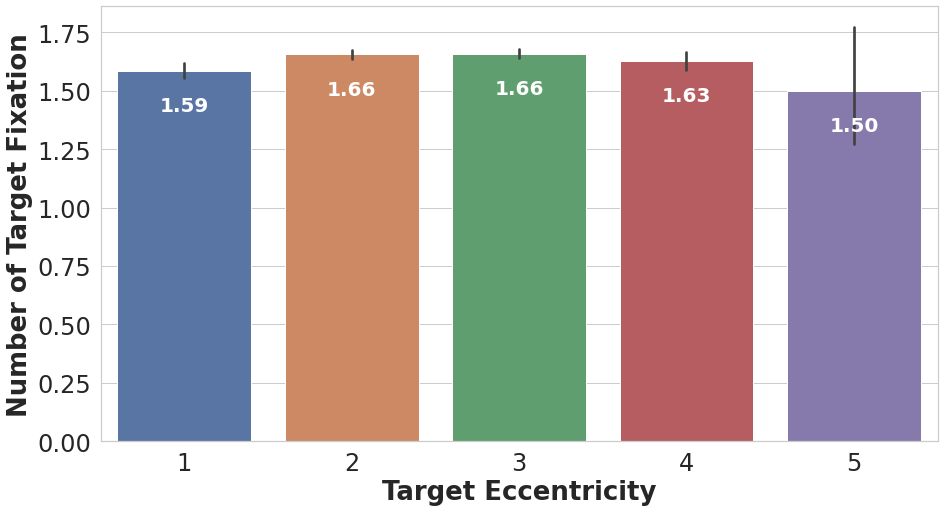}\label{fig:eccent_1_single}
\caption{Search performance versus target eccentricity (target's euclidean distance from the center of an image). (a) Average reaction time. (b) Average fixation number. (c) Average non-target fixation duration. (d) Average target fixation duration. (e) Average number of target fixation.}
\label{fig:distance}
\end{figure*}

Another crucial factor in determining the search performance is the eccentricity of the target objects, i.e. the distance of the target from the center of the image. It has been believed that humans tend to look toward the central parts of images. Also, during trials the participants initially placed their gaze on a dot in the center of images. We therefore assume that if a target is at the center of the image, participants should be able to find it more rapidly. We test our hypothesis by plotting the previous variables but this time versus the targets eccentricity. We compute the euclidean distance between the center pixel of the target's bounding box and the center pixel of the image. We then divide the distances into 6 bins, each containing 50 pixel euclidean distance. Bin 0 corresponds to 0-50 pixel distance, bin 1 contains 50-100 pixel distance, ..., and bin 5 contains 250-300 pixel distance. The results can be seen in figure \ref{fig:distance}

The results show that the average reaction time, average number of fixations and average distractor fixation duration increase as the target's distance from the center increases. This validates our previous assumption that targets further away from the center are harder to find with more fixations and longer response time. The observers also fixate longer at non-target objects while looking for the distant targets and therefore have higher distractors fixation duration. The number of fixations made on the target and the duration of these fixations are almost constant across various levels of targets' eccentricity.

\subsection{Error Rate}

\begin{figure*}[ht!]
\centering
\includegraphics[width=0.45\linewidth]{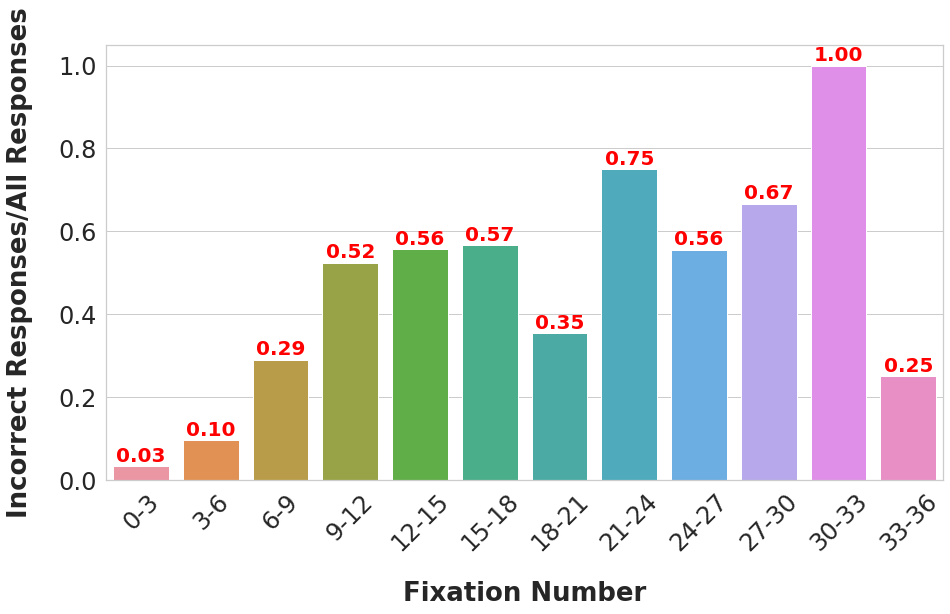}\label{fig:incorr_1_single}
\hfill
\includegraphics[width=0.45\linewidth]{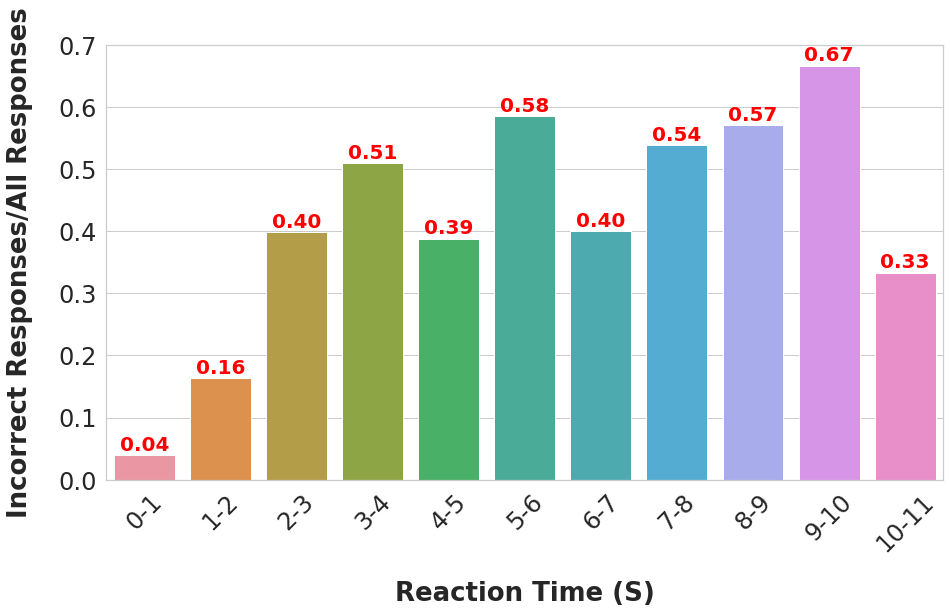}\label{fig:incorr_2_single}
\includegraphics[width=0.45\linewidth]{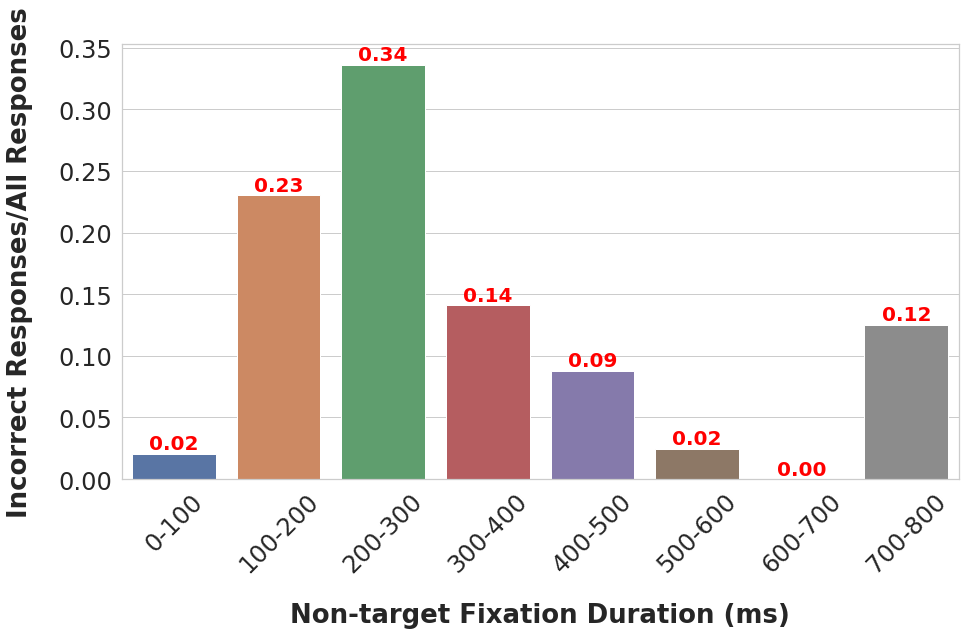}\label{fig:incorr_3_single}
\hfill
\includegraphics[width=0.45\linewidth]{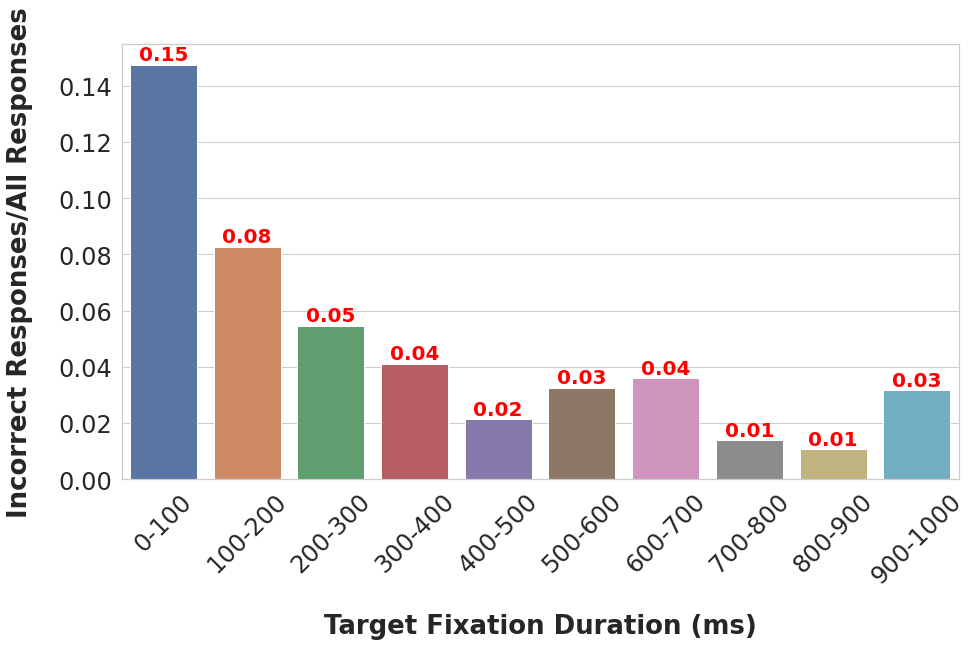}\label{fig:incorr_4_single}
\caption{Ratio of the incorrect responses to the total number of responses versus: (a) Reaction time. (b) Fixation number. (c) Non-target fixation duration. (d) Target fixation duration.}
\label{fig:incorrect_re}
\end{figure*}

Finally, we investigate the correlation between the error rate of participants with average reaction time, average number of fixations, average non-target fixation duration, and average target fixation duration, in figure \ref{fig:incorrect_re}. The error rate is defined as the number of incorrect responses (by all participants) divided by the total number of responses for each bin of the histograms. 

From figure \ref{fig:incorrect_re}, we can derive that longer reaction times correspond to higher error rates. As in more difficult search images that the target is less detectable, participants spend longer time searching the images, and some eventually fail to find the target. This explanation does not contradict our earlier discussion on individual differences in reaction time. We previously stated that participants who have longer average reaction time had less number of incorrect responses, as a result of their more careful search. However, trials with higher reaction times contain both types of participants: those who perform exhaustive search and those who perform non-exhaustive search. Thus, the individual differences are not the focus of our argument here.
The same pattern exists for the number of fixations. A higher number of fixation typically corresponds to a more difficult search, and consequently higher error rates. Furthermore, fixating at non-target objects for a duration of 200-300 milliseconds correspond to the highest erroneous responses. Also, shorter fixations at target objects correspond to more incorrect answers. Thus we can infer that when participants recognize the target, they fixate at it longer than when they look at the target but do not recognize it.

\section{Discussion and Conclusion}

A potential application of these results could be in visual marketing, i.e. using visual information to guide customers' attention. By extending these analyses to a 3D environment, we can direct customers' attention toward healthier products. For instance, our data analyses show that an increased target eccentricity decreases the distraction (attention to non-target objects) and makes the target search easier. Same principal also applies in marketing strategies. In \cite{Wedel2008EyeTF}, Wedel suggests that ``a larger number of facings, and top and/or central placement of products on the shelves increases attention to and consideration of the brand \cite{Wedel2008EyeTF}.'' Hence to encourage healthy products, one could place them at the top and central levels of the shelf which are more aligned with humans' average height.

Another factor that facilitates the visual search is the bigger size of the target, which causes more visibility. Similarly, Wedel suggests that in print advertising and banners, the size of the ad has a positive effect on attention. Hence by choosing the neighbor products of a healthy product such that the healthy one is more differentiating (e.g. slightly bigger) and visible, we might increase the shoppers' chance of buying healthy products.

Another important component in encouraging healthy products is by directing customers' attention to the health warnings and nutrition labels on the products' package. For instance, Wedel discusses that information near the center of the label, which is typically dense, is less visible compared to the bottom or top of the label. Also, by using colors and pictures, health warnings could be designed more noticeable. \cite{Wedel2008EyeTF}


\bibliographystyle{unsrt}  
\bibliography{references}

\end{document}